\pdfoutput=1

\documentclass[11pt]{article}

\usepackage{EMNLP2022}

\usepackage{times}
\usepackage{latexsym}

\usepackage[T1]{fontenc}

\usepackage[utf8]{inputenc}

\usepackage{microtype}
\usepackage{amsmath, amssymb}
\usepackage{booktabs}
\usepackage{graphicx}
\usepackage{multirow}
\usepackage{float, subfig}
\usepackage{xspace}

\title{\textsc{Hpt}: Hierarchy-aware Prompt Tuning for  Hierarchical Text Classification}

\newcommand{\modelname}{\textsc{Hpt}\xspace}

\author{
Zihan Wang$^1$$\footnotemark[2]$ \quad Peiyi Wang$^1$$\footnotemark[2]$ \quad Tianyu Liu$^2$ \quad Binghuai Lin$^2$ \\ \textbf{Yunbo Cao$^2$ \quad Zhifang Sui$^1$ \quad Houfeng Wang$^1$$\footnotemark[1]$}
\\ 
$^1$ MOE Key Laboratory of Computational Linguistics, Peking University, China \\
$^2$ Tencent Cloud Xiaowei \\
 {\{wangzh9969, wangpeiyi9979\}@gmail.com}; {\{szf, wanghf\}@pku.edu.cn}\\
 {\{rogertyliu, binghuailin, yunbocao\}@tencent.com;} 
}

\begin{document}

\maketitle

\renewcommand{\thefootnote}{\fnsymbol{footnote}}

\begin{abstract}
Hierarchical text classification (HTC) is a challenging subtask of multi-label classification due to its complex label hierarchy.
Recently, the pretrained language models (PLM)
have been widely adopted in HTC through a fine-tuning paradigm.
However, in this paradigm, there exists a huge gap between the classification tasks with sophisticated label hierarchy and the masked language model (MLM) pretraining tasks of PLMs and thus the potentials of PLMs can not be fully tapped.
To bridge the gap, in this paper, we propose \textbf{\modelname}, a \underline{\textbf{H}}ierarchy-aware \underline{\textbf{P}}rompt \underline{\textbf{T}}uning method to handle HTC from a multi-label MLM perspective.
Specifically, we construct a dynamic virtual template and label words that take the form of soft prompts to fuse the label hierarchy knowledge and introduce a zero-bounded multi-label cross entropy loss to harmonize the objectives of HTC and MLM.
Extensive experiments show \modelname achieves state-of-the-art performances on $3$ popular HTC datasets and is adept at handling the imbalance and low resource situations. Our code is available at \url{https://github.com/wzh9969/HPT}.

\end{abstract}

\footnotetext[2]{Equal contribution.}
\footnotetext[1]{Corresponding author.}

\renewcommand{\thefootnote}{\arabic{footnote}}

\section{Introduction}


Hierarchical text classification (HTC) aims to categorize a text into a set of labels with a structured class hierarchy (commonly modeled as a tree) \cite{silla2011survey}.
HTC is a multi-label text classification problem, where the classification result corresponds to one or more paths of the hierarchy  \cite{zhou2020hierarchy}.
The major challenge of HTC is to model the large-scale, imbalanced, and structured label hierarchy \cite{mao2019hierarchical}.

\begin{figure}[t]
    \centering
    \includegraphics[width=0.9\linewidth]{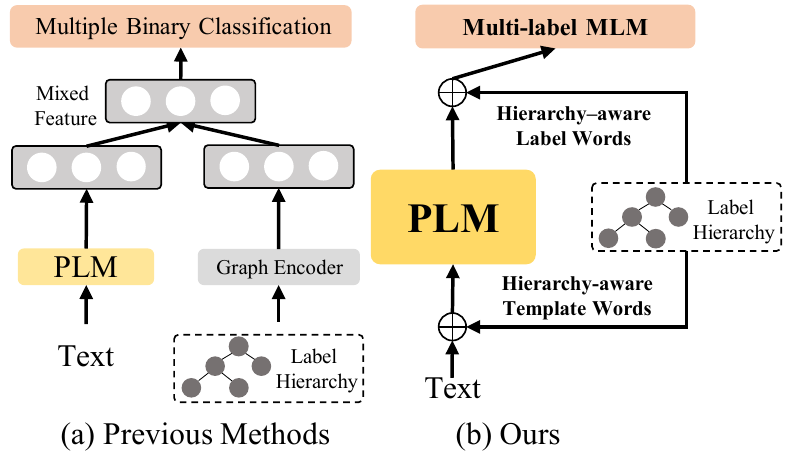}
    \caption{Comparison of previous methods and our \modelname. (a) Previous  models formulate HTC as a multiple binary classification problem, and utilize the PLM in a fine tuning paradigm . (b) \modelname follows a prompt tuning paradigm that transforms HTC into a hierarchy-aware multi-label MLM problems. }
    \label{fig:intro}
\end{figure}

As shown in Figure \ref{fig:intro}(a), existing state-of-the-art HTC models \cite{zhou2020hierarchy,deng2021htcinfomax,chenhierarchy,zhao2021hierarchical}  separately extract text and label hierarchy features by utilizing text and graph encoders, and then fuse the two sources of features into a final representation for text classification.
Specifically, \citet{chenhierarchy} takes the advantages of powerful pretrained language models (PLMs) in HTC through a fine-tuning paradigm, where they use PLMs as the text encoder.
In this paradigm, the PLMs are trained to inference with complex label hierarchy.

Despite the success of the fine-tuning paradigm, some recent studies suggest that it may suffer from distinct training strategies in the pretraining and fine-tuning stages, which restrains the finetuned models to take full advantage of knowledge in PLMs \cite{chen2021knowprompt}.
Therefore, a new paradigm known as \textit{prompt tuning} is proposed to bridge the gap between the downstream tasks and the pretraining tasks of PLMs, which can tap the full potential of PLMs.
By warping the text 
(e.g., ``\textbf{x}'') 
into the model input 
(e.g., ``\textbf{x} is \texttt{[MASK]}'' )
and taming the PLMs to complete the masked cloze test,
prompt tuning has achieved promising performances on the flat text classification where labels have no hierarchy \cite{shin2020eliciting}.

How about the performances of the prompt tuning in HTC?
In the pilot study, we test flat prompt tuning methods on HTC and surprisingly find that they are even comparable with the state-of-the-art models in HTC. 
This result suggests that the expressive power of PLMs has been undermined in the prior HTC methods due to the pretraining-finetuning gap.
Although the flat prompt tuning methods have somewhat narrowed the gap, there still remain two challenges while combining PLMs with HTC.
\begin{enumerate}
    \item \textbf{hierarchy and flat gap}. 
Labels of HTC lie on a sophisticated hierarchy while MLM pretraining and flat prompt tuning do not take label hierarchy into consideration.
\item \textbf{multi-label and multi-class gap}.
HTC is a multi-label classification problem where the output labels are interconnected with a hierarchy while MLM pretraining is formulated as multi-class classification.
\end{enumerate}

To bridge these two gaps, as shown in Figure \ref{fig:intro}(b), we propose a hierarchy-aware prompt tuning (\modelname) method that solves HTC from a multi-label MLM perspective.
In detail, to bridge the hierarchy and flat gap,  
we incorporate the label hierarchy knowledge into soft prompts with continuous representation. 
Specifically, we incorporate the depth and width information in the label hierarchy into different virtual template words, which is helpful to alleviate the label imbalance problem as verified by our experiments. 
To bridge the multi-label and multi-class gap, we transform HTC into a multi-label MLM problem by a zero bounded multi-label cross entropy loss which continually seeks to increase the score of the correct label
and decrease the score of the incorrect labels.

We summarize our contributions as follows: 
\begin{itemize}
    \item We propose a hierarchy-aware prompt tuning (\modelname) method for hierarchical text classification. To the best of our knowledge, this is the first investigation on flat and hierarchical prompt tuning in HTC. 
\item We summarize two challenging gaps between HTC and masked language modeling (MLM). To bridge these gaps, we transform HTC into a hierarchy-aware multi-label MLM problem.
\item Extensive experiments demonstrate that our proposed model achieves the new state-of-the-art results on three popular datasets, and is adept at handling label imbalance and low resource situations.
\end{itemize}

\section{Related Work}
\subsection{Hierarchical Text Classification}
Hierarchical text classification (HTC) is a challenge task due to its large-scale, imbalanced, and structured label hierarchy \cite{mao2019hierarchical}.
Existing work for HTC could be categorized into local and global approaches based on their ways of utilizing the label hierarchy \cite{zhou2020hierarchy}: local approaches build classifiers for each node or level while the global ones build only one classifier for the entire graph. Although early works on HTC mainly focus on local approaches \cite{wehrmann2018hierarchical,shimura2018hft,banerjee2019hierarchical}, global approaches soon become mainstream.
The early global approaches neglect the hierarchical structure of labels and view the problem as a flat multi-label classification \cite{johnson2014effective}. Later on, some work try to coalesce the label structure by meta-learning \cite{wu2019learning}, reinforcement learning \cite{mao2019hierarchical}, and attention module \cite{zhang2020hcn}. Although such methods can capture the hierarchical information, \citet{zhou2020hierarchy} demonstrate that encode the holistic label structure directly by a structure encoder can further improve performance. Following this research, a bunch of models tries to study how the hierarchy should interact with the text. Both \citet{chen2020hyperbolic} and \citet{chenhierarchy} embed word and label hierarchy jointly in a same space. \citet{deng2021htcinfomax} constrains label representation with information maximization. \citet{zhao2021hierarchical} designs a self-adaption fusion strategy to extract features from text and label. \citet{wang2022incorporating} adopts contrastive learning to directly inject the hierarchical knowledge into text encoder.


\subsection{Prompt tuning}
\label{sec:prompt}
Prompt tuning \cite{schick2021exploiting} aims to transform the downstream NLP task into the pretraining task of the pretrained language models (PLM), which can bridge their gap and better utilize PLM.
The most popular pretraining task of PLM is MLM \cite{devlin2018bert}, which masks some words in the input text and requires PLM to recover these masked words.
The prompt tuning methods can be broadly divided into $2$ categories:
(1) \textit{Hard prompt} \cite{gao2020making,schick2021exploiting}.
The hard prompt methods select template and label words from the vocabulary of PLM, which require carefully manual designing.
(2) \textit{Soft prompt} \cite{hambardzumyan-etal-2021-warp,qin2021learning}. Soft prompt methods first create some continuous vectors as template and label embeddings, and then find the best prompt using the training examples, which eliminate the need for manually-designed prompts.

\section{Preliminaries}
\subsection{Problem Definition}
For each hierarchical text classification (HTC) dataset, we have a predefined label hierarchy $\mathcal{H}=(\mathcal{Y}, E)$, where $\mathcal{Y}$ is the label set (also the node set of $\mathcal{H}$) and $E$ is the edge set.
In HTC, given an input text \textbf{x}, the models aim to categorise it into a label set $Y \subseteq \mathcal{Y}$.
Specifically, we focus on a setting where every node except the root has one and only one father so that the hierarchy can be simplified as a tree-like structure. In this case, labels can be organized into layers where labels in the same layer have the same depth in the tree.
The predicted label set $Y$ corresponds to one or more paths in $\mathcal{H}$.

\subsection{Vanilla Fine Tuning for HTC}
Given an input text \textbf{x}, the vanilla Fine Tuning method first converts it  to ``\texttt{[CLS]} \textbf{x} \texttt{[SEP]}'' as the model input, and then utilizes the PLM to encode it.
After that, it utilizes $\mathbf{h_{\texttt{[CLS]}}}$, the hidden state of ``\texttt{[CLS]}'', to predict the labels of the input text.
Previous methods \cite{chenhierarchy, wang2022incorporating} based on the PLM all follow this fine tuning paradigm.

\subsection{Prompt Tuning for HTC}
\label{sec:pt}
To bridge the gap between the pretraining task and the downstream tasks, prompt tuning has been proposed.
We adopt $2$ typical flat text classification prompt methods to HTC.
\paragraph{Hard Prompt} For a text ``\textbf{x}'', hard prompt first applies a template and fills the input into it. For HTC, we choose ``\texttt{[CLS]} \textbf{x} \texttt{[SEP]} The text is about \texttt{[MASK]} \texttt{[SEP]}'' as template. The PLM is then asked to predict the ``\texttt{[MASK]}'' slot, which outputs a score for every word in the vocabulary. A verbalizer is then selected for each label to represent its meaning: the score of filling that verbalizer into the ``\texttt{[MASK]}'' slot is the prediction score on according label. We select the head word (the root word on the dependency tree) of the label name as verbalizer to represent according label.

\paragraph{Soft Prompt} For a text ``\textbf{x}'', soft prompt append a fixed number of learnable virtual template words to the text (i.e., ``\texttt{[CLS]} \textbf{x} \texttt{[SEP]} \texttt{[V1]} \texttt{[V2]} ... \texttt{[V8]} \texttt{[MASK]} \texttt{[SEP]}'' in case of $8$) as template. During training, the PLM learns to predict the ``\texttt{[MASK]}'' slot as well as tunes virtual template words.
For HTC, we create a learnable label embedding as verbalizer for each hierarchical label.

\bigskip
Since HTC is a multi-label classification problem, following previous works, 
both the vanilla fine tuning and $2$ typical prompt tuning methods finally conduct a multiple binary classification. The output of PLM is normalized by sigmoid instead of the original softmax to predict on each label and the loss function is changed to binary cross entropy.

Although we can modify these $2$ typical prompt tuning methods for HTC, the essence of this challenge has not been considered. As mentioned, the existing prompt methods experience two major gaps when migrating to HTC:
\begin{enumerate}
    \item \textbf{Hierarchy and flat gap.} Both soft prompt and hard prompt do not take labels into account until prediction, and PLM views all candidate words as equal. 
    Previous works suggest that incorporating label dependency instead of modeling them as flat classification is essential for alleviating the label imbalance \cite{gopal2013recursive}.
    \item \textbf{Multi-label and multi-class gap.} Previous works on HTC view the problem as multiple binary classification but MLM is designed for multi-class classification. Prompting aims to bridge the gap between pretraining and fine-tuning but the gap still exists if we use the sigmoid normalization and binary cross entropy loss functions for HTC during fine-tuning.
\end{enumerate}

\section{Methodology}
\begin{figure*}[t]
    \centering
    \includegraphics[width=\linewidth]{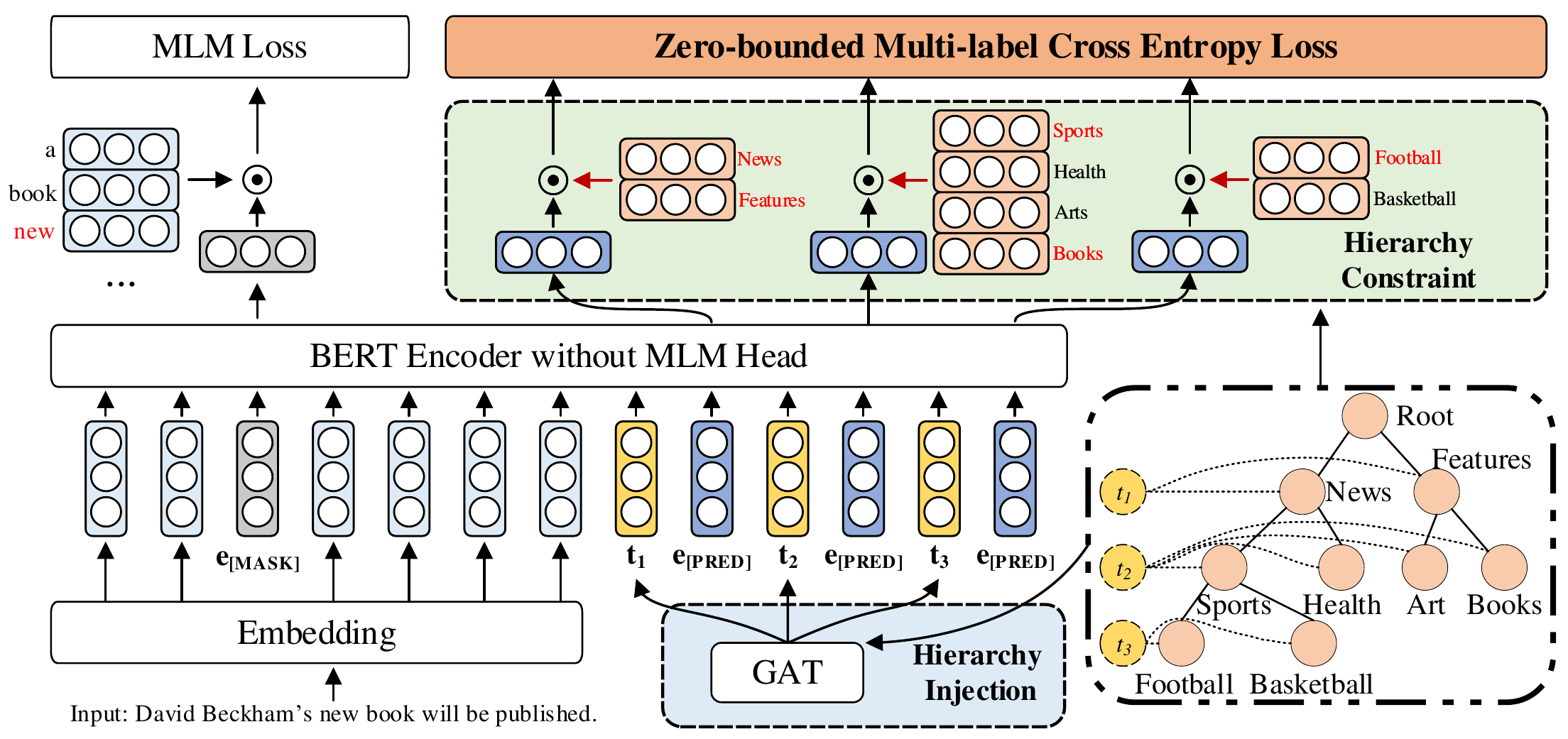}
    \caption{The architecture of \modelname during training. \modelname  transforms HTC into a hierarchy-aware multi-label MLM problem that focuses on bridging \textit{two} gaps between HTC and MLM. (1) To bridge the hierarchy and flat gap, \modelname incorporates the label hierarchy knowledge into dynamic virtual template and label words construction. (2) To bridge the multi-label and multi-class gap. \modelname transforms HTC into a multi-label MLM task with a zero-bounded multi-label cross entropy loss.}
    \label{fig:overview}
\end{figure*}

In this section, we introduce a hierarchy-aware prompt tuning method to solve HTC from a multi-label MLM perspective. 

\subsection{Hierarchy-aware Prompt}
To bridge the \textit{hierarchy and flat gap},
we create the prompt with the label hierarchy constraint and injection.

\subsubsection{Hierarchy Constraint}
To incorporate the label hierarchy, we propose a layer-wise prompt. Since the label hierarchy is a tree, we construct templates based on the depth of the hierarchy.
Given a predefined label hierarchy $\mathcal{H}=(\mathcal{Y},E)$ with a depth of $L$ and input text \textbf{x}, the template is ``\texttt{[CLS]} \textbf{x} \texttt{[SEP]} \texttt{[V1]} \texttt{[PRED]} \texttt{[V2]} \texttt{[PRED]} ... \texttt{[VL]} \texttt{[PRED]} \texttt{[SEP]}''. Instead of a fixed number of template words as soft prompt, we have a dynamic template which has template words (from \texttt{[V1]} to \texttt{[VL]}) the same number as hierarchy layers. We use a special \texttt{[PRED]} token for label prediction, indicating a multi-label predication.

We use BERT \cite{devlin2018bert} as text encoder, which first embeds input tokens to embedding space:
\begin{equation}
    \mathbf{T}=[\mathbf{x}_1, \dots, \mathbf{x}_N,\mathbf{t}_1,\mathbf{e}_P,\dots,\mathbf{t}_L,\mathbf{e}_P]
\label{eq:input}
\end{equation}
where $\mathbf{X}=[\mathbf{x}_1, \dots, \mathbf{x}_N]$ is word embeddings and $\mathbf{e}_P$ is the embedding of \texttt{[PRED]}, which is initialized by the \texttt{[MASK]} token of BERT. $\{\mathbf{\mathbf{t}_i}\}_{i=1}^L$ are layer-wise template embeddings. Similar to soft prompt, template embeddings are randomly initialized and are learned through training. Here we omit special tokens of BERT (\texttt{[CLS]} and \texttt{[SEP]}) for clarity.

BERT then encodes $\mathbf{T}$ to achieve the hidden states:
\begin{equation}
    \mathbf{H}=[\mathbf{h}_1,\dots,  \mathbf{h}_N,\mathbf{h}_{t_1},\mathbf{h}_P^1,\dots,\mathbf{h}_{t_L},\mathbf{h}_P^L]
\end{equation}
where $\mathbf{h}_P^i$ is the hidden state of the $i$-th $\mathbf{e}_P$, which corresponds to the $i$-th layer of the label hierarchy.

For verbalizer, we create a learnable virtual label word $v_i$ for each label $y_i$ and initialize its embedding $\mathbf{v}_i$ with the averaging embedding of its corresponding tokens. Instead of predicting all labels in one slot, as shown in the green part of Figure \ref{fig:overview}, we divide labels into different groups according to their layers and constrain \texttt{[PRED]} to only predict labels on one layer. To this end, each template word \texttt{[Vi]} is followed by a \texttt{[PRED]} token for predictions on the $i$-th layer. By splitting predictions into different slots, the model may learn better about the dependency between labels across different layers and somewhat solve the label imbalance.

Formally, for $\mathbf{h}_P^m$, we define a verbalizer ${\rm Verb}_m$ as follows:
\begin{equation}
    \begin{aligned}
        {\rm Verb}_m(y_i) =\left\{
        \begin{aligned}
         & v_i, &  y_i \in \mathcal{N}_m \\
         & \varnothing, &  {\rm Others} \\
        \end{aligned}
        \right. 
    \end{aligned}
\end{equation}
where $\mathcal{N}_m$ is the label set of the $m$-th layer and $\varnothing$ denotes that there is no label word for labels at other layers.

\subsubsection{Hierarchy Injection}
\label{sec:tem}
The hierarchy constraint only introduces depth of labels but lacks their connectivity. To make full use of the label hierarchy in an MLM-manner, 
we further inject the per-layer label hierarchy knowledge into template embedding.

As shown in the blue part of Figure \ref{fig:overview},
a $K$-layer stacked Graph Attention Network (GAT) \cite{kipf2017semi} is adopted to model the label hierarchy.
Given a node $u$ at the $k$-th GAT layer, the information interaction and aggregation operation is defined as follows:
\begin{equation}
       \mathbf{g}_{u}^{(k + 1)} = \mathrm{ReLU} (\sum_{v\in\mathcal{N}(u) \bigcup \{u\}} \frac{1}{c_{u}} \mathbf{W}^{(k)} \mathbf{g}_{v}^{(k)} )
\end{equation}
where $\mathcal{N}(u)$ denotes the neighbors for node $u$ , $c_u$ is a normalization constant and $\mathbf{W}^{(l)}\in \mathbb{R}^{d_m \times d_m}$ is the trainable parameter.

To achieve per layer knowledge for our layer-wise prompt, we create $L$ virtual nodes $t_1, \dots, t_L$ (colored in yellow) and connect $t_i$ with all label nodes at the $i$-th layer in $\mathcal{H}$. In this way, these virtual nodes can aggregate information from a certain hierarchical level through artificial connections.
For the first GAT layer, we adopt the virtual label word $\mathbf{v_i}$ for node $y_i \in \mathcal{Y}$ as its node feature and assign template embedding $\mathbf{\mathbf{t}_i}$ to virtual node $t_i$ as its node feature.

GAT is then applied to the new graph and it outputs representations $\mathbf{g}_{t_i}^K$ for virtual node $t_i$, which has gathered knowledge from the $i$-th layer. We utilize a residual connection to achieve the $i$-th graph template embedding:
\begin{equation}\label{eq:5}
   \mathbf{t'}_i = \mathbf{t}_i + \mathbf{g}_{t_i}^K
\end{equation}
where the new template embedding with hierarchy knowledge, $\mathbf{t'}_i$, is injected into BERT replacing $\mathbf{t}_i$ in Equation \ref{eq:input}.

\subsection{Zero-bounded Multi-label Cross Entropy Loss} \label{sec:zmcel}
Since hierarchical text classification is a multi-label classification problem,
previous methods \cite{zhou2020hierarchy,chenhierarchy,zhao2021hierarchical} mainly regard HTC as a multiple binary classification problem and utilize the binary cross entropy (BCE) as their loss function:
\begin{equation}
    \mathcal{L}_{BCE} = -\sum_{i}^C (y_{i}{\rm log}(s_{y_i}) + (1- y_i){\rm log}(1-s_{y_i}))
    \label{Eq:bce}
\end{equation}


where $s_{y_i}$ is the predicted sigmoid score of the label $y_i$ for the input. 
As illustrate in Equation \ref{Eq:bce}, BCE ignores the correlation between labels.
In contrast, the masked language modeling
is a multi-class classification task, which is optimized with the cross entropy (CE) loss:
\begin{equation}\label{Eq:ce}
 \begin{split}
    \mathcal{L}_{CE} &= -{\rm log}\frac{e^{s_{y_t}}}{\sum_{i=1}^Ce^{s_{y_i}}} \\& = {\rm log}(1 + \sum_{i=1, i\neq t}^Ce^{s_{y_i} - s_{y_t}})
\end{split}
\end{equation}
where $y_t$ is the gold label for the input.
As shown in Equation \ref{Eq:ce}, CE forces the score of the gold label is greater than all other labels, which directly models the label correlation.

To harmonize their objectives and bridge this \textit{multi-label and multi-class gap},
in this paper, instead of calculating the score of each label separately, we expect the scores of all target labels are greater than all non-target labels. We use a multi-label cross entropy (MLCE) loss \cite{sun2020circle, kexuefm-7359}:
\begin{equation}\label{Eq:cl}
    \mathcal{L}_{MLCE} =  {\rm log}(1 + \sum_{y_i \in \mathcal{N}^{n}}\sum_{ y_j \in \mathcal{N}^{p}}e^{s_{y_i} - s_{y_j}})
\end{equation}
where $\mathcal{N}^{p}$ and $\mathcal{N}^{n}$ are the target and non-target label set of the input text.

However, Equation \ref{Eq:cl} is impracticable since we cannot know the number of target labels during inference even if the positive (target) labels and negative (other) labels are separated.
To fix this glitch, following \citet{kexuefm-7359}, we introduce an anchor label with a constant score $0$ in MLCE and hope that the scores of the target labels and the non-target labels are all greater and less than $0$ respectively. Thus, we form a zero-bounded multi-label cross entropy (ZMLCE) loss:

\begin{equation}
    \begin{split}
    &\mathcal{L}_{ZMLCE} = {\rm log}(1 + \sum_{y_i \in \mathcal{N}^{n}}\sum_{ y_j \in \mathcal{N}^{p}}e^{s_{y_i} - s_{y_j}} \\
    &+\sum_{y_i \in \mathcal{N}^{n}}e^{s_{y_i}-0} +  \sum_{ y_j \in \mathcal{N}^{p}}e^{0 - s_{y_j}}) \\
    &= {\rm log}(1+\sum_{y_i \in \mathcal{N}^{n}}e^{s_{y_i}}) + {\rm log}(1+\sum_{y_i \in \mathcal{N}^{p}}e^{-s_{y_i}}) \\
    \end{split}
    \label{eq:ZMLCE}
\end{equation}

To be consistent with the hierarchy constraint, we adopt ZMLCE at each label hierarchy layer for the layer-wise prediction.
Formally, for the $m$-th layer with scores predicted by $\mathbf{h}_P^m$, we add layer constraints as follow:
\begin{equation}
\begin{split}
  \mathcal{L}_{ZMLCE}^m = & {\rm log}(1+\sum_{y_i \in \mathcal{N}^{n}_m}e^{s_{y_i}}) 
  \\& + {\rm log}(1+\sum_{y_i \in \mathcal{N}^{p}_m}e^{-s_{y_i}})
\end{split}
\end{equation}
where $s_{y_i}=\mathbf{v}_i^T\mathbf{h}_P^m + b_{im}$ and $b_{im}$ is a learnable bias term.
$\mathcal{N}_m^{p}$ and $\mathcal{N}_m^{n}$ are the target and non-target label set at the $m$-th layer for the input text respectively.

We keep the original MLM loss as BERT pretraining and the final loss $\mathcal{L}_{all}$ is the sum of ZMLCE losses at different layers and the MLM loss:
\begin{equation}
    \mathcal{L}_{all}= \sum_{m=1}^L\mathcal{L}_{ZMLCE}^m + \mathcal{L}_{MLM}
\end{equation}
We randomly mask $15$\% words of the text to compute the MLM loss $\mathcal{L}_{MLM}$.
During inference, we select labels with scores greater than $0$ as our prediction.
A comparison between our method and existing prompt methods is in Appendix \ref{sec:app3}.


\section{Experiments}
\subsection{Experiment Setup}

\paragraph{Datasets and Evaluation Metrics}
We experiment on Web-of-Science (WOS) \cite{kowsari2017hdltex}, NYTimes (NYT) \cite{sandhaus2008new}, and RCV1-V2 \cite{lewis2004rcv1} datasets for analysis. 
The statistic details are illustrated in Table \ref{tab:1}.
We follow the data processing of previous work \cite{zhou2020hierarchy,chenhierarchy} and measure the experimental results with Macro-F1 and Micro-F1. 

\paragraph{Baselines}
For systematic comparisons, we introduce a variety of hierarchical text classification baselines and compare \modelname with two typical prompt learning methods.
1) \textbf{TextRCNN} \cite{lai2015recurrent}. A simple network of bidirectional GRU followed by CNN. It is a traditional text classification model adopted by HiAGM, HTCInfoMax, and HiMatch as their text encoder.
2) \textbf{BERT} \cite{devlin2018bert}. A widely used pretrained language model that can serve as a text encoder. Among previous work, only HiMatch introduces BERT as text encoder so that we implement other baselines with BERT replaced.
3) \textbf{HiAGM} \cite{zhou2020hierarchy}. HiAGM exploits the prior probability of label dependencies through Graph Convolution Network and applies soft attention over text feature and label feature for the mixed feature.
4) \textbf{HTCInfoMax} \cite{deng2021htcinfomax}. HTCInfoMax improves HiAGM by maximizing text-label mutual information and matching the label feature to a prior distribution.
5) \textbf{HiMatch} \cite{chenhierarchy}. HiMatch views the problem as a semantic matching problem and matches the relationship between the text semantics and the label semantics.
6) \textbf{HGCLR} \cite{wang2022incorporating}. HGCLR regulates BERT representation by contrastive learning and introduces a new graph encoder.

\paragraph{Implement Details} We implement our model using PyTorch with an end-to-end fashion. Following previous work \cite{chenhierarchy}, we use \texttt{bert-base-uncased} as our base architecture. We use a single layer of GAT for hierarchy injection. The batch size is set to $16$. The optimizer is Adam with a learning rate of $3e^{-5}$. We train the model with train set and evaluate on development set after every epoch and stop training if the Macro-F1 does not increase for $5$ epochs. All of the hyperparameters have not been tuned. For baseline models, we follow the hyperparameter tuning procedure in their original paper. We use a length of $8$ template words for soft prompt in accordance with \modelname.

\subsection{Main Results}
\label{sec: main_r}
Table \ref{tab:2} illustrates our main results. 
As is shown, ``HardPrompt'' and ``SoftPrompt'' outperform the vanilla fine tuning BERT on all $3$ datasets and achieve a comparable result with the state-of-the-art method on RCV1-V2.
This result shows the superiority of the prompt tuning paradigm since it adapts HTC to BERT to some extent.

By bridging the gaps between HTC and MLM, our \modelname achieves new state-of-the-art results on all $3$ datasets.
Comparing to HiMatch \cite{chenhierarchy}, our model introduces no extra parameter so that these improvements demonstrate that \modelname can better utilize the pretrained language model.
Although HGCLR \cite{wang2022incorporating} introduces a new graph encoder, our model achieves consistent improvements on all datasets with a simple GAT.
In addition, the depths of the label hierarchy for WOS, RCV1-V2, and NYT are $2$, $4$, and $8$ respectively, which can reflect the respective difficulty of the label hierarchy.
\modelname outperforms both the baseline BERT and HGCLR by increasing margins on WOS, RCV1-V2, and NYT respectively, showing that hierarchy-aware prompt can better handle more difficult label hierarchy.


\begin{table*}[t]
\centering
\resizebox{\linewidth}{!}{
\begin{tabular}{lcccccc}
\toprule
\multirow{2}{*}{Model}                                & \multicolumn{2}{c}{WOS (Depth 2)} & \multicolumn{2}{c}{RCV1-V2 (Depth 4)} & \multicolumn{2}{c}{NYT (Depth 8)} \\ \cmidrule(l){2-7} 
                                                      & Micro-F1        & Macro-F1        & Micro-F1          & Macro-F1          & Micro-F1        & Macro-F1       \\ \midrule
TextRCNN \cite{zhou2020hierarchy}    & 83.55           & 76.99           & 81.57             & 59.25             & 70.83           & 56.18          \\
HiAGM \cite{zhou2020hierarchy}       & 85.82           & 80.28           & 83.96             & 63.35             & 74.97           & 60.83          \\
HTCInfoMax \cite{deng2021htcinfomax} & 85.58           & 80.05           & 83.51             & 62.71             & -               & -              \\
HiMatch \cite{chenhierarchy}         & 86.20           & 80.53           & 84.73             & 64.11             & -               & -              \\ \midrule
BERT \cite{wang2022incorporating}            & 85.63      & 79.07      & 85.65        & 67.02      & 78.24      & 66.08        \\
BERT+HiAGM\cite{wang2022incorporating}                                 & 86.04           & 80.19           & 85.58             & 67.93             & 78.64           & 66.76          \\
BERT+HTCInfoMax\cite{wang2022incorporating}                            & 86.30           & 79.97           & 85.53             & 67.09             & 78.75               & 67.31              \\
BERT+HiMatch \cite{chenhierarchy}    & 86.70           & 81.06           & 86.33             & 68.66             & -               & -              \\
 HGCLR \cite{wang2022incorporating}   & 87.11           & 81.20           & 86.49           & 68.31            & 78.86              & 67.96              \\
\midrule
BERT+HardPrompt (Ours)                                  & 86.39           & 80.43           & 86.78             & 68.78             & 79.45           & 67.99          \\
BERT+SoftPrompt (Ours)                                     & 86.57           & 80.75           & 86.53             & 68.34             & 78.95           & 68.21          \\
\modelname (Ours)           & \textbf{87.16}           & \textbf{81.93}           & \textbf{87.26}             & \textbf{69.53}         
                                                        & \textbf{80.42}           & \textbf{70.42}          \\ \bottomrule
\end{tabular}
}
\caption{F1 scores on $3$ datasets. Best results are in boldface.
}

\label{tab:2}
\end{table*}

\begin{table}[t]
\centering
\resizebox{\linewidth}{!}{
\begin{tabular}{lcc}
\toprule
Ablation Models   & Micro-F1 & Macro-F1 \\ \midrule

\modelname             & \textbf{80.49}    & \textbf{71.07}    \\
\textit{r.m.} hierarchy constraint          &   80.32  &   70.58  \\
\textit{r.m.} hierarchy injection           &  80.41   &   69.71  \\
\textit{r.p.} BCE loss & 79.74    & 70.40    \\
\textit{r.m.} MLM loss          &   80.16  &   70.78  \\
with random connection          & 80.12    &  69.42 \\

\bottomrule
\end{tabular}
}
\caption{Performance when remove some components of \modelname on the development set of NYT. \textit{r.m.} stands for \textit{remove}. \textit{r.p.} stands for \textit{replaced with}.}
\label{tab:ablation}

\end{table}




\subsection{Ablation Study}
To illustrate the effect of our proposed mechanisms, we conduct ablation studies by removing one component of our model at a time. We test on NYT dataset in this and following sections because it has the most complicated label hierarchy and it can better demonstrate how our method reacts to the hierarchy.

After removing the hierarchy constraint, the template has only one \texttt{[PRED]} token and the model needs to recover all label words according to its hidden state.
As shown in Table \ref{tab:ablation}, the Micro-F1 and Macro-F1 drop slightly, which shows the effectiveness of our layer-wise prompt. 
By removing the hierarchy injection (i.e., remove Equation \ref{eq:5}), the model cannot access the connectivity of the label hierarchy and drops $1.36$ on Macro-F1. From this decline, we can see that the hierarchy injection is essential for the performance of labels with few instances. By incorporating an extra structural encoder, the model can learn label features from training instances from other classes based on the hierarchical dependencies between them. As a result, the hierarchy injection significantly boosts the performance of scarce classes.
At last, both the performances of using BCE loss instead of ZMLCE loss (\textit{r.p.} BCE loss) and removing MLM loss (\textit{r.m.} MLM loss) drop, which shows it is important to bridge the gap of optimizing objectives between HTC and MLM.

To further illustrate the effectiveness of the hierarchy injection, we test our model with random connection. As a reminder, during hierarchy injection, we connect virtual nodes with according labels with the same depth. Random connection adds random connections based on that connection. For each label, it connects the label to another virtual node randomly.

As in the last row of Table \ref{tab:ablation}, the variant with random connection drops over 1\% on Macro-F1 score. This result illustrates that connections that violate the label hierarchy have adverse effects. The destructiveness of a contradicting input like random connection even outweighs removing the hierarchy completely (\textit{r.m.} hierarchy injection), reflecting that the proposed \modelname indeed gains instructive information from the label hierarchy. More discussions on the connection of virtual nodes are elaborate in Appendix \ref{app:1}. Ablation results on other datasets are in Appendix \ref{sec:app4}.
%









\subsection{Interpreting on Representation Space}
In this section, we hope to intuitively show how the label hierarchy is incorporated and what the prompt has learned.
The virtual label words are learned in the same space as word embedding, so they can be interpreted by their similarities with meaningful words.
Therefore, 
we illustrate the top $8$ nearest words of $2$ labels in the NYT dataset, \textit{National Hockey League} (NHL) and \textit{News and Features} (NF).
As shown in table \ref{tab:case}, despite some meaningless words, the model indeed learns some interpretable features. For NHL, the label words of \modelname consist of the semantic of \textit{football}, which is the brother node of Hockey (the father node of NHL) in the label hierarchy.
For NF, the label words of \modelname consist of the semantic of \textit{theatre}, which is the father node of NF.
After removing the hierarchy knowledge (r.m. hierarchy), these semantics disappear from label words of NHL and NF.
These results intuitively show \modelname incorporates the hierarchy knowledge into the pretrained language model and bridges the gap between HTC and MLM. 

\begin{table}[t]
\centering
\small
\resizebox{1\linewidth}{!}{
\setlength{\tabcolsep}{2pt}{
\begin{tabular}{@{}cllll@{}}
\toprule
                        & \multicolumn{4}{c}{Top 8 nearest words}                                                     \\ \cmidrule(l){2-5} 
\multirow{-2}{*}{\begin{tabular}[c]{@{}c@{}}Label\\ (different layers separated by `/')\end{tabular}} & \multicolumn{2}{c}{\modelname}    & \multicolumn{2}{c}{\modelname (r.m. hierarchy)} \\ \midrule
                        & [1] hockey   & [2] league   & [1] hockey   & [2] national   \\
                        & [3] national & [4] 2011   & [3] league      & [4] 2012          \\
                        & [5] 2013     & [6] \#\#\textasciicircum{}  & [5] 2008   & [6] 1996          \\
          \multirow{-4}{*}{\begin{tabular}[l]{@{}l@{}}News/Sports/Hockey/\\ National Hockey League\end{tabular}}              & [7] 2012     & [8] {\color{red} football} & [7] 2010        & [8] 2014          \\
           \midrule
                    & [1] features & [2] .       & [1] .     & [2] features      \\
                    & [3] and      & [4] the     & [3] and   & [4] the           \\
                    & {[5] \color{red}theatre}   & [6] ;     & [5] ,    & [6] ;         \\
     \multirow{-4}{*}{\begin{tabular}[l]{@{}l@{}}Features/Theater/\\ News and Features\end{tabular}}                & [7] ,        & [8] news   & [7] of    & [8] news     \\
\bottomrule
\end{tabular}
}}
\caption{Top $8$ nearest words of $2$ learnable virtual label words in NYT dataset. 
}
\label{tab:case}
\end{table}

\subsection{Results on Imbalanced Hierarchy}
\begin{figure}[t]
    \centering
    \subfloat[]{
    \label{fig:layers:a}
        \includegraphics[width=0.5\linewidth]{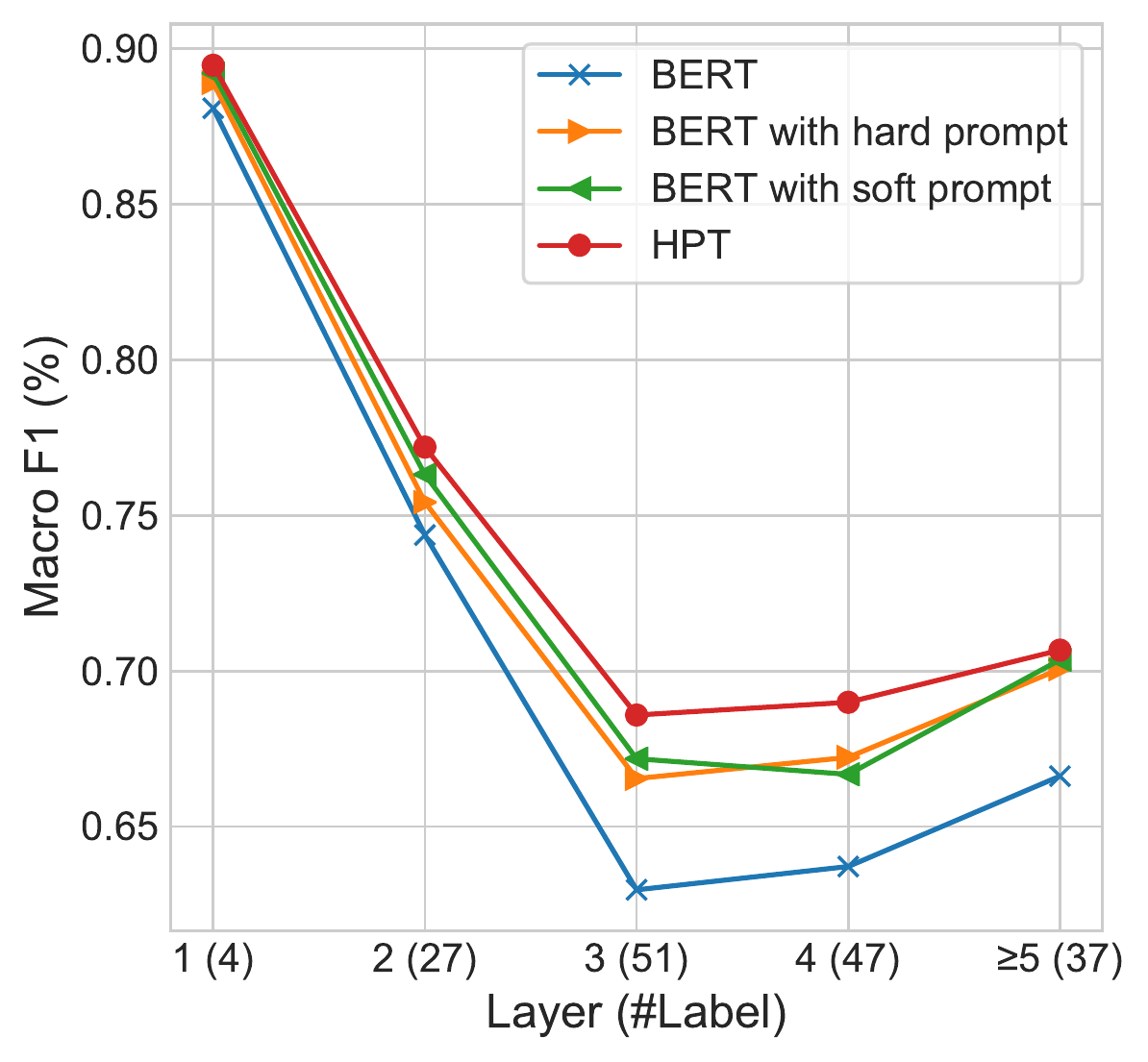}
    }
    \subfloat[]{
    \label{fig:layers:b}
        \includegraphics[width=0.5\linewidth]{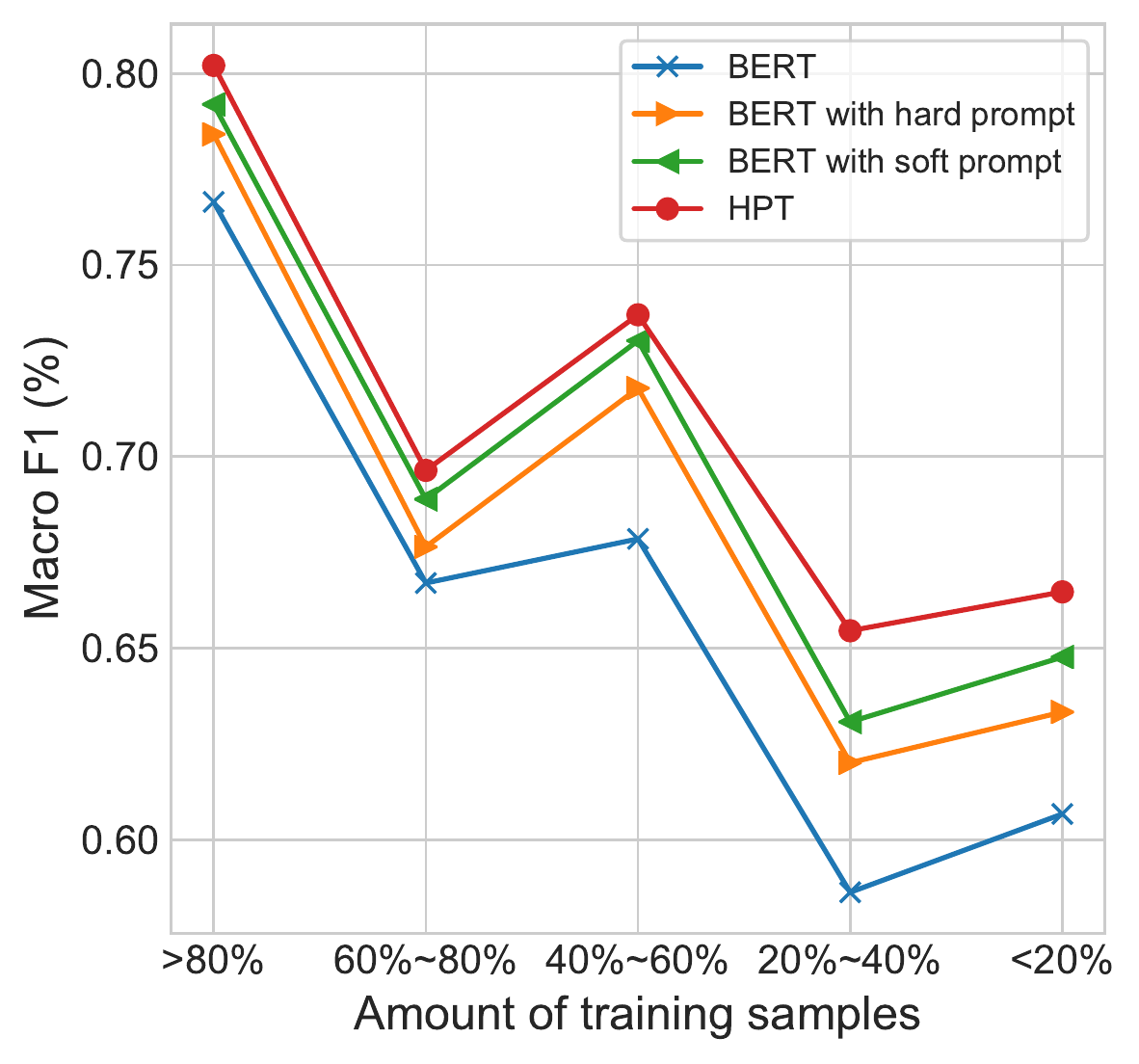}
    }
    \caption{Macro F1 scores of label clusters on the development set of NYT. (a) Label clusters grouped by depth in the hierarchy. 
    (b) Label clusters grouped by amount of training samples. >80\% represents cluster of top 20\% labels ranking by amount of training samples. The rest clusters are arranged similarly.
    }
    \label{fig:layers}
\end{figure}
One of the key challenges of hierarchical text classification is the imbalanced label hierarchy. 
In this section, we analyze how our model resolves the issue of imbalance on the development set of NYT.

For HTC, the imbalance can be viewed from two perspectives. For one, number of labels at different depths of the hierarchy is imbalanced. As shown in Figure \ref{fig:layers:a}, medium layers (depth 3 and 4) have more labels than deep or shallow layers, where all models have poor performances. Comparing with other baselines, \modelname mainly boost the performance of medium levels.
For another, instance of each label is various. Take NYT dataset as an example, the ratio of the maximum and minimum amount of training samples of a label is over $100$. In Figure \ref{fig:layers:b}, we cluster labels into $5$ bunches depending on their amounts of training samples. Our model largely improves the performance of labels with few training instances, showing that our method can alleviate the long-tail effect to some extent.


\begin{figure}[t]
    \centering
    \includegraphics[width=\linewidth]{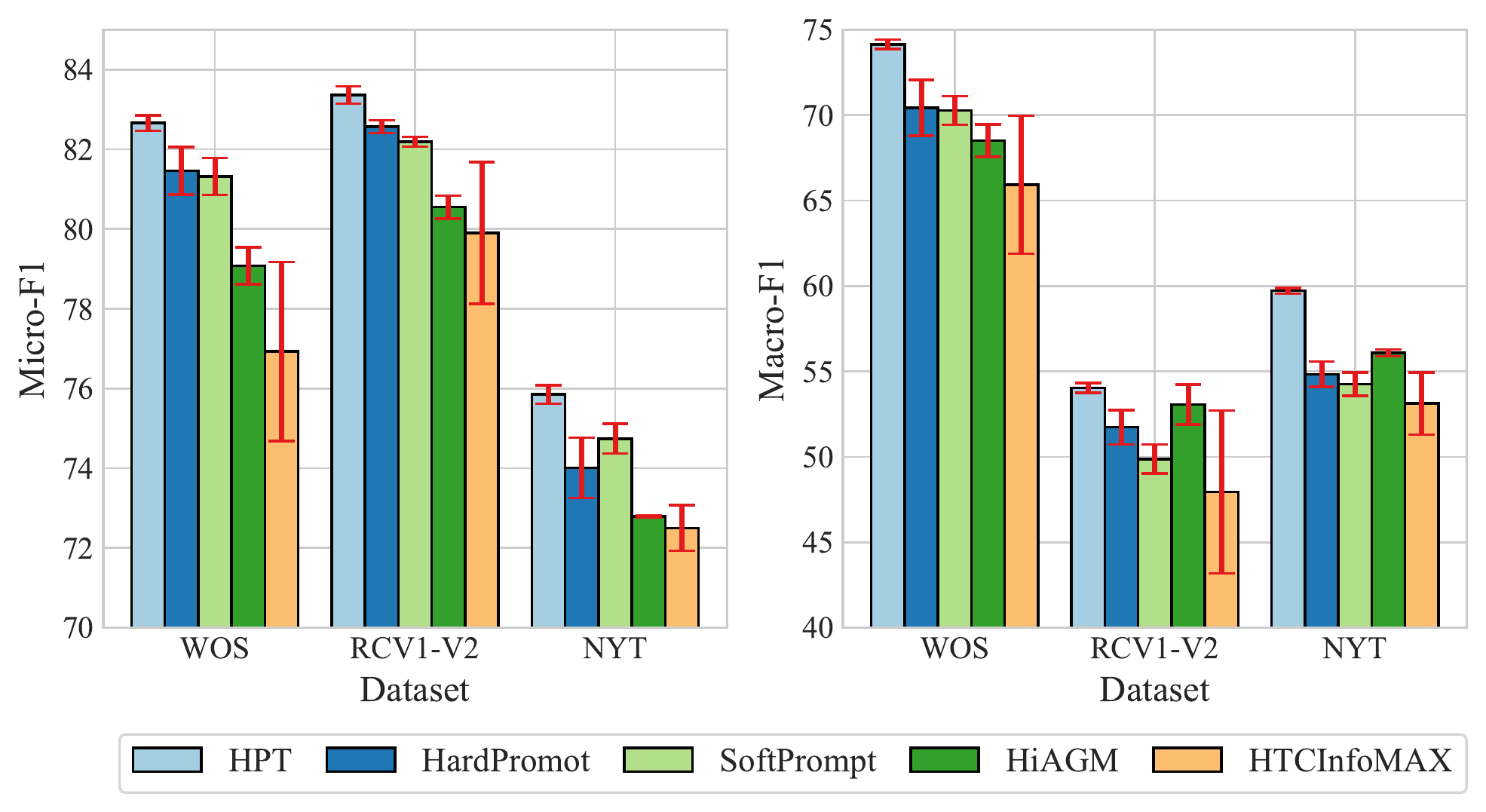}
    \caption{F1 scores on $3$ mini training dataset with only $10$\% training instances of the full training dataset. We report the average scores with standard deviation over $3$ different runs.
    }
    \label{fig:low}
\end{figure}

\subsection{Results on Low Resource Setting}
To further evaluate the potential of our method, we conduct experiments in low-resource settings. Since the problem is multi-label, the commonly used N-way K-shot setting is hard to define so we simply sample $10\%$ of training data.
As previous HTC works do not consider the low resource setting (LRS), we reproduce baseline models in LRS. Besides less training data, other settings follow the main experiment.

The comparison of LRS experimental results is shown in Figure \ref{fig:low}.
Among baseline methods, prompt-based models outperform non-prompt-based models on $3$ datasets, which shows the advantages of prompt methods in LRS.
Our model outperforms all baseline models and has better stability (lower standard deviation) on all $3$ LRS datasets.
Comparing with the full resource setting (FRS) (i.e., main results), the performance gap between \modelname and other baselines increases on the LRS.
For example, on RCV1-V2, \modelname outperforms ``BERT+HTCinfoMAX'' $2.13$ and $6.09$ Macro-F1 scores in FRS and LRS, respectively, which shows the potential of our method.



\section{Conclusion}
In this paper, we propose a hierarchy-aware prompt tuning (\modelname) method to bridge the gaps between HTC and MLM.
To bridge the \textit{hierarchy and flat gap}, \modelname incorporates the label hierarchy knowledge into virtual template and label words.
To bridge the \textit{multi-label and multi-class gap}, \modelname introduces a zero-bounded multi-label cross entropy loss to harmonize the objectives of HTC and MLM.
\modelname transforms HTC into a hierarchy-aware multi-label MLM task, which can better tap the potential of the pretrained language model in HTC.
Extensive experiments show that our method achieves state-of-the-art performances on $3$ popular HTC datasets, and is adept at handling the imbalance and low resource situations.

\section*{Limitations}
Prompting methods needs pretrained language model as backbone. Our work is based on the masked language model (MLM) task but it is not a universal component of PLM. As a result, our approach is only applicable to PLMs which incorporate MLM. Despite such limited choices, comparing to other HTC works which adopt PLM as a replaceable text encoder, our approach takes more advantage of PLMs by considering how they are trained. Another limitation is the constraint of maximum sequence length. Although the length limitation of PLM is extensively existed, our approach needs extra tokens for template, and that further shortens the length of input text. Even so, the experiment results indicate that our method performs better than the raw PLM so that this sacrifice is worthy. Notice that the length of our template is proportional to the depth of the label hierarchy, so \modelname may fail to datasets with extreme hierarchy depth.

\section*{Acknowledgements}
We thank all the anonymous reviewers for their constructive feedback.
The work is supported by National Natural Science Foundation of China under Grant No.62036001, PKU-Baidu Fund (No. 2020BD021) and NSFC project U19A2065.

\bibliographystyle{acl_natbib}
\bibliography{acl}
\appendix
\section{Data Statistics}
\begin{table}[ht]
\resizebox{\linewidth}{!}{
\begin{tabular}{ccccccc}
\toprule
Dataset & $|Y|$   & Depth &   Avg($|y_i|$)   & Train  & Dev   & Test    \\ \midrule
WOS     & 141 & 2     & 2.0  & 30,070 & 7,518 & 9,397   \\
NYT     & 166 & 8     & 7.6  & 23,345 & 5,834 & 7,292   \\
RCV1-V2    & 103 & 4     & 3.24 & 20,833 & 2,316 & 781,265 \\ \bottomrule
\end{tabular}
}
\caption{Data statistics. $|Y|$ is the number of classes. Depth is the maximum level of hierarchy. Avg($|y_i|$) is the average number of classes per sample.}
\label{tab:1}
\end{table}

\section{Example of Different Prompt Methods} \label{sec:app3}

We provide some detailed examples here to explain the difference between our \modelname with existing prompt methods.

Templates of hard prompt, soft prompt and \modelname are illustrated in Table \ref{tab:template}. \textbf{x} is the original text and \texttt{[CLS]} and \texttt{[SEP]} are special tokens of BERT. \texttt{[V1]} to \texttt{[VN]} in soft prompt are $N$ virtual template words which are learnable embeddings, and the number $N$ is predefined. Our method has $L$ virtual template words. They are output embeddings of graph encoder as in Equation \ref{eq:5} and $L$ is the number of hierarchy layers. 
Our method uses a special token \texttt{[PRED]} for multi-label prediction (Section \ref{sec:zmcel}), whereas hard and soft prompt use the same \texttt{[MASK]} token as BERT, which is proposed for single-label predictions.

\begin{table}[h]
\resizebox{\linewidth}{!}{
\begin{tabular}{ccccccc}
\toprule
Method & Template   \\ \midrule
Hard prompt     & \begin{tabular}{c} \texttt{[CLS]} \textbf{x} \texttt{[SEP]} The text is \\ about \texttt{[MASK]} \texttt{[SEP]} \end{tabular}\\ \midrule
Soft prompt     & \begin{tabular}{c}  \texttt{[CLS]} \textbf{x} \texttt{[SEP]} \texttt{[V1]} \\ \texttt{[V2]} ... \texttt{[VN]} \texttt{[MASK]} \texttt{[SEP]} \end{tabular} \\ \midrule
\modelname    & \begin{tabular}{c}  \texttt{[CLS]} \textbf{x} \texttt{[SEP]} \texttt{[V1]}  \texttt{[PRED]} \texttt{[V2]} \\ \texttt{[PRED]} ... \texttt{[VL]} \texttt{[PRED]} \texttt{[SEP]}  \end{tabular} \\ \bottomrule
\end{tabular}
}
\caption{Example templates of hard prompt, soft prompt and our method. \textbf{x} is the original text.}
\label{tab:template}
\end{table}

\section{Discussion on Different Connections of Hierarchy Injection}\label{app:1}
\begin{figure}[t]
    \centering
    \subfloat[Depth increasing]{
    \label{fig:conn:a}
        \includegraphics[width=0.4\linewidth]{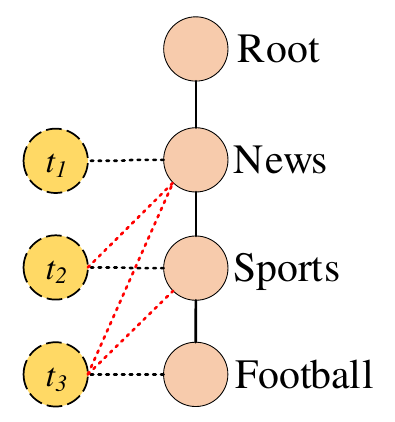}
    }
    \subfloat[Random connection]{
    \label{fig:conn:b}
        \includegraphics[width=0.4\linewidth]{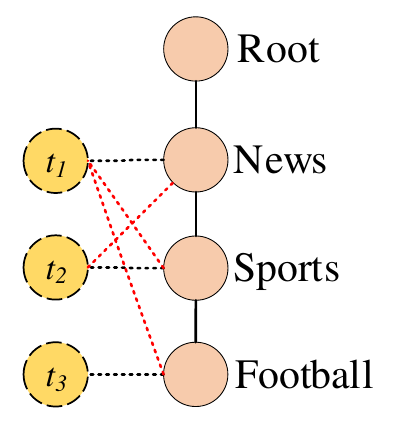}
    }
    \caption{Two connections to aggregate node features. They add more connections (red dash line) besides the original connections (black dash line) (a) Depth increasing connects a virtual node with labels on the same and shallower layers. 
    (b) Random connection adds random connection per node.
    }
    \label{fig:conn}
\end{figure}

During hierarchy injection, we connect virtual nodes with according labels with same depth, but this connection is not unique. Besides random connection, we further test our model with a variant. Depth increasing connects a virtual nodes with labels on the same and shallower layers, i.e., virtual node $t_i$ connects with all label nodes on 1st to $i$-th layers. Figure \ref{fig:conn} is an illustration of theses two connections.

As in the third row of Table \ref{tab:conn}, variant with depth increasing behaves similarly to the original one. This observation illustrates that the impact of the connection of virtual nodes is not significant as long as it contains logical hierarchical information. Comparing with random connection which violates the label hierarchy and has adverse effects, this result reflects that the proposed \modelname is aware of the label hierarchy on the secondary side.

\begin{table}[t]
\centering
\resizebox{0.95\linewidth}{!}{
\begin{tabular}{lcc}
\toprule
Ablation Models   & Micro-F1 & Macro-F1 \\ \midrule

\modelname             & \textbf{80.49}    & \textbf{71.07}    \\
\textit{r.m.} hierarchy injection           &  80.41   &   69.71  \\
with depth increasing           &  80.48   &   70.95  \\
with random connection          & 80.12    &  69.42 \\

\bottomrule
\end{tabular}
}
\caption{Performance of different connections of hierarchy injection on the development set of NYT. \textit{r.m.} stands for \textit{remove}.}
\label{tab:conn}
\end{table}

\section{Ablation results on WebOfScience and RCV1-V2} \label{sec:app4}
\begin{table}[h]
\centering
\resizebox{\linewidth}{!}{
\begin{tabular}{lcc}
\toprule
Ablation Models   & Micro-F1 & Macro-F1 \\ \midrule
\modelname             & \textbf{87.88}    & \textbf{81.68}    \\
\textit{r.m.} hierarchy constraint          &   87.34  &   81.27  \\
\textit{r.m.} hierarchy injection           &  87.58   &   81.54  \\
\textit{r.p.} BCE loss & 87.17    & 80.78    \\
\textit{r.m.} MLM loss          &   87.22  &   81.36  \\
with random connection          & 87.56    &  81.42 \\

\bottomrule
\end{tabular}
}
\caption{Performance when remove some components of \modelname on the development set of WOS. \textit{r.m.} stands for \textit{remove}. \textit{r.p.} stands for \textit{replaced with}.}
\label{tab:ablation1}
\end{table}

\begin{table}[h]
\centering
\resizebox{\linewidth}{!}{
\begin{tabular}{lcc}
\toprule
Ablation Models   & Micro-F1 & Macro-F1 \\ \midrule
\modelname             & \textbf{88.37}    & \textbf{70.12}    \\
\textit{r.m.} hierarchy constraint          &   87.62  &   69.04  \\
\textit{r.m.} hierarchy injection           &  87.57  &   68.53  \\
\textit{r.p.} BCE loss & 87.79    & 68.12    \\
\textit{r.m.} MLM loss          &   87.83  &   69.76  \\
with random connection          & 88.22    &  68.86 \\

\bottomrule
\end{tabular}
}
\caption{Performance when remove some components of \modelname on the development set of RCV1-V2. \textit{r.m.} stands for \textit{remove}. \textit{r.p.} stands for \textit{replaced with}.}
\label{tab:ablation2}
\end{table}

The hierarchy of WOS dataset only has two layer so that structural information of WOS is weak. So, in Table \ref{tab:ablation1}, removing or disturbing such information have little influence.

After replacing ZMLCE loss with BCE loss, Macro-F1 decreases dramatically on all datasets. Although BCE loss indeed can solve the multi-label problem, ZMLCE loss is a better choice theoretically and experimentally.

\end{document}